\documentclass[runningheads]{llncs}
\usepackage{graphicx}
\usepackage{microtype}
\usepackage{graphicx}
\usepackage{subfigure}
\usepackage{booktabs} 
\usepackage{amsfonts}

\usepackage{amsmath}
\usepackage{placeins}

\begin{document}
\title{Multi-Path Learnable Wavelet Neural Network for Image Classification}

%
%
\author{D.D.N. De Silva \and H.W.M.K. Vithanage \and K.S.D. Fernando \and I.T.S. Piyatilake.}

\authorrunning{De Silva \textit{et al.}}
%
\institute{Department of Computational Mathematics, University of Moratuwa, Moratuwa 10400, Sri Lanka.}
\maketitle              
\begin{abstract}

Despite the remarkable success of deep learning in pattern recognition, deep network models face the problem of training a large number of parameters. In this paper, we propose and evaluate a novel multi-path wavelet neural network architecture for image classification with far less number of trainable parameters. The model architecture consists of a multi-path layout with several levels of wavelet decompositions performed in parallel followed by fully connected layers. These decomposition operations comprise wavelet neurons with learnable parameters, which are updated during the training phase using the back-propagation algorithm. We evaluate the performance of the introduced network using common image datasets without data augmentation except for SVHN and compare the results with influential deep learning models. Our findings support the possibility of reducing the number of parameters significantly in deep neural networks without compromising its accuracy.

\end{abstract}

\section{Introduction}
\label{introduction}

Over the past decade, deep learning has become the state of the art in pattern recognition. Deep learning belongs to the set of machine learning techniques, which utilize artificial neural networks. Inspired by the structure of the human brain, these networks allow learning the representations of data at multiple levels of abstractions. Neural networks that consists of multiple hidden layers are known as deep neural networks and they are the driving force of deep learning. These hidden layers contain a massive number of parameters and they help the network to learn complex patterns in data. This is one of the fundamental reasons that makes deep learning an enormous success. Typical deep neural networks contain a large number of interconnections between neurons.  These connections are represented by tunable parameters and enable capturing representations that are more abstracts. During the training phase, Backpropagation algorithm \cite{cite_40} updates the parameters of the hidden layers in order to minimize the error of the final network output. Moreover, the complexity of neural networks is often expressed as a function of the number of parameters. There are notable examples to prove above claim such as AlexNet \cite{cite_1}, VGG19 \cite{cite_2} and ResNet \cite{cite_25} which contain approximately $60$ million, $140$ million and $60$ million parameters respectively.

However, tuning this sheer number of parameters demand massive sets of training data and consumes a significant amount of computational power. This limits applicability of deep learning and a trade-off must be made between computational resources and performance of the network. A neural network that can deliver the same level of performance with a lesser number of parameters can be introduced as a desirable solution.

In this research work, we propose and evaluate a wavelet artificial neural network for pattern recognition with a far less number of learnable parameters. This network is capable of generating classification outputs by using hierarchical feature representations developed by the $2$D Discrete Wavelet Transform (DWT). Each level of DWT decomposes input vision data and develops suitable feature representations. We parameterize the wavelet transform by including learnable parameters, which are learned during the training. Compared to conventional deep learning models, our architecture benefits by having a far less number of parameters to be tuned. The core architecture we propose is a combination of a set of wavelet transform pipelines and fully connected layers. We have empirically evaluated different versions of the proposed network by using a preliminary dataset and then we selected the best performing architecture and it was tested for multiple datasets. The results are compared with several baseline deep learning architectures that have influenced the field.

\section{Related Work}
\label{related_work}

The wavelet transform is a powerful tool for processing data and developing time-frequency representations. A thorough theoretical background on wavelets is explained in \cite{cite_16,cite_17}. Applying wavelet transform in the context of neural networks is not novel. Earlier work \cite{cite_3,cite_4} has presented a theoretical approach for wavelet-based feed-forward neural networks. The ability to use wavelet based interpolation for real time unknown function approximation has been researched by Bernard \textit{et al.} \cite{cite_5}. In this case, the results have been achieved with a relatively less number of coefficients due to the high compression ability of the wavelets. The work by Alexandridis \textit{et al.} \cite{cite_6} has proposed a statistical model identification framework in applying wavelet networks and it is investigated under many subjects including architecture, initialization, variable and model selection. Literature indicates applications of wavelet based neural networks in many different fields such as signal classification and compression \cite{cite_8,cite_7,cite_9}, in time series predicting \cite{cite_10,cite_11,cite_12}, electrical load forecasting \cite{cite_14,cite_13} and power distribution recognition \cite{cite_15}.

In terms of deep learning, wavelet-based methods have been used in many computer vision applications through Convolutional Neural Networks (CNN) due its ability of diverse information extraction. Texture classification using wavelet CNN \cite{cite_28}, multi-scale face super-resolution \cite{cite_29}, image super-resolution \cite{cite_30} and edge feature enhancing \cite{cite_31} are notable applications. A multi-level wavelet CNN model for image restoration has been introduced by Liu \textit{et al.} \cite{cite_27}. Cotter \textit{et al.} \cite{cite_18} proposed a novel layer that performs convolution filtering and activation in wavelet domain and then returns to pixel space. We have included the results from this work in the section \ref{results} in order to compare it with our findings. Furthermore, a deep wavelet network has been developed by combining several wavelet networks whose transfer function is based on wavelets and autoencoders in \cite{cite_48}. Oyallon \textit{et al.} \cite{cite_20} have developed a hybrid wavelet deep learning network based on wavelets scattering transform \cite{cite_19}. This has introduced a simple classifier, which was later improved by Singh \textit{et al.} \cite{cite_21}. Savareh \textit{et al.} \cite{cite_22} have proposed a wavelet enhanced fully convolutional network architecture for brain tumor segmentation and DWT has been combined with the neural network classifier \cite{cite_26} for the same application. Moreover, the work by Williams \textit{et al.} \cite{cite_47} preprocesses the data in th wavelet domain before feeding to a CNN to attain greater accuracies. We have also compared the results from this work with ours. An encoder-decoder CNN architecture with DWT at the encoder and inverse DWT at the decoder to reduce the spatial resolution and increase the receptive field for dense pixel wise prediction has been introduced by Ma \textit{et al.} \cite{cite_23}. Moreover, a wavelet neural network has been presented in \cite{cite_24} for the purpose of voice and noise separation.

Although a significant amount of work has been done in this area, limited attention has been paid on parameterizing the wavelet transform in the context of neural networks. Therefore, we introduce and evaluate a learnable wavelet neural network system to fill the above mentioned gap while retaining the advantage of having a far less number of parameters compared to conventional deep learning architectures.

\section{Approach}

\subsection{Artificial Neural Networks}

Artificial neural network is an information processing paradigm vaguely inspired by the biological neural networks that constitute animal brains. They are not preprogrammed with any algorithm but can act as a framework for many machine learning algorithms to process complex data inputs. Currently, there are multiple architectures of neural networks specializing in many different machine learning domains. Convolutional neural networks are frequently applied in image processing while recurrent architectures such as Long-short term memory (LSTM) and Gated recurrent units (GRU) are used for language processing tasks. However, common to all of these architectures we can find a neural network to be composed of highly interconnected processing modules. These interconnections enable the network to extract complex features in data. For example, in deep learning, first layers of the neural network extract simple features followed by next layers extracting more abstract feature representations which are finally mapped to the output. Machine learning systems based on deep neural networks have the ability to do automatic feature extraction without human intervention and capability to approximate complex non-linear functions to remarkable levels of accuracy. These abilities have made deep neural networks to be applicable in a vast area of applications. This includes applications such as natural language processing, bioinformatics, computer vision, and speech recognition.

\subsection{Discrete Wavelet Transform}

The Wavelet transform is a series expansion of signals (well suited for transient signals). It closely resembles the way that the Fourier series uses a combination of sinusoids to represent a signal in discrete form, while a wavelet is an oscillating finite energy signal of time or space. Furthermore, this transform provides a multi-resolution analysis of signals (including images) localized in both time and frequency or scale. Expansion of a signal using a set of different scales of resolution coefficients enables representing a wide range of details. In other words, narrow wavelets allow capturing high resolution or greater details. Furthermore, lower resolution coefficients can be derived from the higher resolution coefficients and in DWT this is known as the tree-structured filter banks decomposition. The multi-resolution analysis can be mathematically represented using the summation of approximated version $V$, and detailed versions $W$ of the decomposed signal as defined below.
\begin{equation}\label{eq.1}
V_{j} = W_{j+1} \oplus V_{j+1} = W_{j+1} \oplus W_{j+2} \oplus \cdots \oplus W_{j+n} \oplus V_{j+n},
\end{equation}
where $j$ and $n$ are scale and the decomposition level respectively. The multi-resolution formulation in DWT employs two basis functions: wavelet function $\psi_{j,k}(t)$ and scaling function $\varphi_k(t)$. The wavelet function is defined as,
\begin{equation}\label{eq.2}
\psi_{j,k}(t) = 2^{j/2}\psi(2^{j}t-k),	\quad\quad\quad	j,k \in \mathbb{Z}
\end{equation}
where $j$ and $k$ parameterize the scale and translation respectively. This parameterization derives from the mother wavelet $\psi(t)$ . The factor $2^{j/2}$ keeps a constant norm independent of scale $j$, and $2^j$ provides the dyadic scaling capability for the dilation. Translation $k$ of the basis function $\varphi(t)$ defines the scaling function as follows:
\begin{equation}\label{eq.3}
\varphi_{k}(t) = \varphi(t-k),	\quad\quad\quad	k \in \mathbb{Z}.
\end{equation}
A wavelet system can be constructed using a set of scaling and wavelet functions to represent any signal or function $f(t)$. Therefore, the function $f(t)$ could be written as,
\begin{equation}\label{eq.4}
f(t) = \sum_{k = -\infty} ^ \infty c(k) \varphi_{k}(t) + \sum_{j = 0} ^ \infty \sum_{k = -\infty} ^ \infty d(j,k) \psi_{j,k}(t),
\end{equation}
where $c(k)$ and $d(j,k)$ are the scaling and wavelet coefficients respectively. These coefficients in the wavelet expansion are called as the DWT of a given signal.

\begin{figure}[ht]
	\begin{center}
		\centerline{\includegraphics[width=120mm]{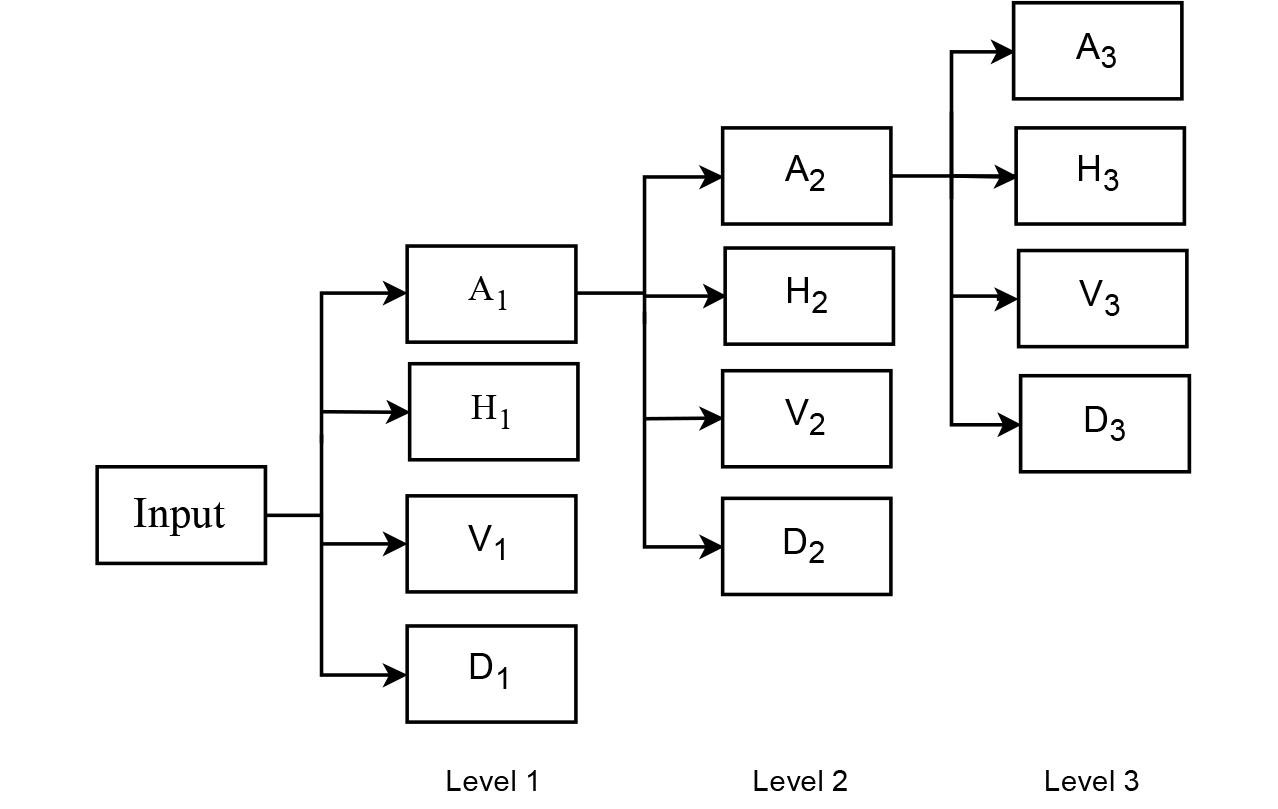}}
		\caption{Discrete wavelet decomposition up to three levels, which outputs one approximation coefficient ($A_i$) and three detail coefficients including horizontal ($H_i$), vertical ($V_i$) and diagonal ($D_i$).}
		\label{dwt_decomposition_tree}
	\end{center}
\end{figure}

In image processing, DWT is used to decompose an image into different levels of frequency interpretations. The DWT is performed by simultaneously passing the image signal through a set of high pass and low pass filters (or filter bank system). The resulting coefficients contain multiple levels of frequency details. The output contains two main types of coefficients: detail coefficients and approximation coefficients (see Figure \ref{dwt_decomposition_tree}). Moreover, one level of decomposition down sample the signal by a factor of two in order to prevent information redundancy. The developed approximation coefficients from one level can be subjected to further DWT to generate more detailed decompositions.

\subsection{Wavelets with Learnable Parameters}

The parameterization of the wavelet based multi-resolution analysis starts with defining the scaling function in (\ref{eq.3}), in the recursive form of:
\begin{equation}\label{eq.5}
\varphi(t) = \sum_n h(n)\sqrt{2}\varphi(2t-n), \quad\quad\quad	n \in \mathbb{Z}
\end{equation}
where the coefficients $h(n)$ are a sequence known as scaling function coefficients or scaling filters. Here, $\sqrt{2}$ maintains the norm of the scaling function. This recursive equation expresses $\varphi(t)$ as a weighted sum of shifted $\varphi(2t)$. For example, the Haar \cite{cite_38} scaling function, simple unit-width, unit height pulse function $\varphi(t)$ can be constructed by $\varphi(2t)$, which satisfies for function coefficients $h(0)=1/\sqrt{2}$ and $h(1)=1/\sqrt{2}$. Therefore, $\varphi(t)$ can be written as
\begin{equation}\label{eq.5.1}
\varphi(t) = \varphi(2t) + \varphi(2t-1).
\end{equation}
According to (\ref{eq.1}) and (\ref{eq.4}), coefficients of wavelet functions at the current level can be derived from the coefficients of the previous level scaling functions. Wavelet function $\psi(t)$ can be represented in the form of weighted sum of shifted $\varphi(2t)$ by
\begin{equation}\label{eq.6}
\psi(t) = \sum_n h_1(n)\sqrt{2}\varphi(2t-n), \quad\quad\quad	
\end{equation}
for the set of wavelet function coefficients $h_1(n)$. Moreover, the orthogonality relationship of wavelet function coefficients to scaling function coefficients is given by
\begin{equation}\label{eq.7}
h_1(n) = (-1)^nh(1-n).
\end{equation}
If $\varphi_{k}(t)$ and $\psi_{j,k}(t)$ are orthonormal, scaling coefficient and wavelet coefficient at the $j^{\mathrm{th}}$ level can be calculated using (\ref{eq.4}) as follows:
\begin{equation}
\begin{aligned} \label{eq.p}
c_j(k) &= \langle f(t), \varphi_{k}(t)\rangle, \enspace \mathrm{and} \\ 
d_j(k) &= \langle f(t), \psi_{j,k}(t)\rangle.
\end{aligned}
\end{equation}
The inner product with the scaling function at a scale of $j+1$ is given by
\begin{equation}\label{eq.10}
c_j(k)=\sum_m h(m-2k) c_{j+1}(m).
\end{equation}
The corresponding relationship for the wavelet coefficient derived from the scaling coefficient at the $j+1$ level is given by
\begin{equation}\label{eq.11}
d_j(k)=\sum_m h_1(m-2k) c_{j+1}(m).
\end{equation}
Therefore, a wavelet system can be designed by choosing the appropriate values for scaling function coefficients $h(n)$. The necessary conditions for $h(n)$ are thoroughly explained in \cite{cite_39}. In this section, we only consider the conditions that are important for designing a parameterized wavelet system. If the solution space of $\varphi(t)$ is finite and if $\int \varphi(t)dt \neq 0$, then
\begin{equation}\label{eq.12}
\sum_n h(n) = \sqrt{2}.
\end{equation}
If integer translations of $\varphi(t)$ are orthogonal as defined by
\begin{equation}\label{eq.13}
\int \varphi(t) \varphi(t-k) dt = E\delta(k)  
= \begin{cases}E, & \mathrm{if}\quad k = 0,\\0 & \mathrm{otherwise.}
\end{cases}
\end{equation}
Then,
\begin{equation}\label{eq.14}
\sum_n h(n)h(n-2k)= \delta(k)  = \begin{cases}1 & \mathrm{if}\quad k = 0, \\ 0 & \mathrm{otherwise.}\end{cases}
\end{equation}
Here $E$ is the energy of the wavelet system. Coefficients $h(n)$, which satisfies the condition in (\ref{eq.14}) are often called as `quadrature mirror filters'. From above conditions we can prove that the norm of $h(n)$ is equal to unity. Therefore,
\begin{equation}\label{eq.15}
\sum_n |h(n)|^2 = 1.
\end{equation}
Based on these necessary conditions we design a wavelet system by choosing the $h(n)$ that gives the best signal decomposition properties. In (\ref{eq.5.1}), a Haar wavelet system can be expressed using length-$2$ $h(n)$ coefficients. The proposed implementation of this research employs length-$6$ $h(n)$ coefficient sequence that satisfies the following necessary conditions:
\begin{equation}\label{eq.16}
\begin{aligned}
h(0) + h(1) + h(2) + h(3) + h(4) + h(5) & = \sqrt2, \\
h^2(0) + h^2(1) + h^2(2) + h^2(3) + h^2(4) + h^2(5) & = 1,  \\
h(0)h(2) + h(2)h(4) + h(1)h(3) + h(3)h(5) & = 0.
\end{aligned}
\end{equation}
As wavelets are wave like finite energy signals, we can solve the system of equations in (\ref{eq.16}) in terms of sinusoidal functions with two defined parameters. Therefore, we define $\alpha$ and $\beta$ to be the arbitrary parameter and the coefficients can be defined as follows:
\begin{equation}\label{eq.17}
\begin{aligned}
h(0) &=[(1+\cos(\alpha)+\sin(\alpha))(1-\cos(\beta)-\sin(\beta))+2\sin(\beta)\cos(\alpha)]/(4\sqrt{2}),   \\
h(1) &=[(1-\cos(\alpha)+\sin(\alpha))(1+\cos(\beta)-\sin(\beta))-2\sin(\beta)\cos(\alpha)]/(4\sqrt{2}),   \\
h(2) &= [1+\cos(\alpha-\beta) + \sin(\alpha-\beta)]/(2\sqrt{2}),   \\
h(3) &= [1+\cos(\alpha-\beta) - \sin(\alpha-\beta)]/(2\sqrt{2}), \\
h(4) &= 1/\sqrt{2} - h(0) - h(2), \\
h(5) &= 1/\sqrt{2} - h(1) - h(3).
\end{aligned}
\end{equation}

\begin{figure}[ht]
	\begin{center}
		\centerline{\includegraphics[width=110mm]{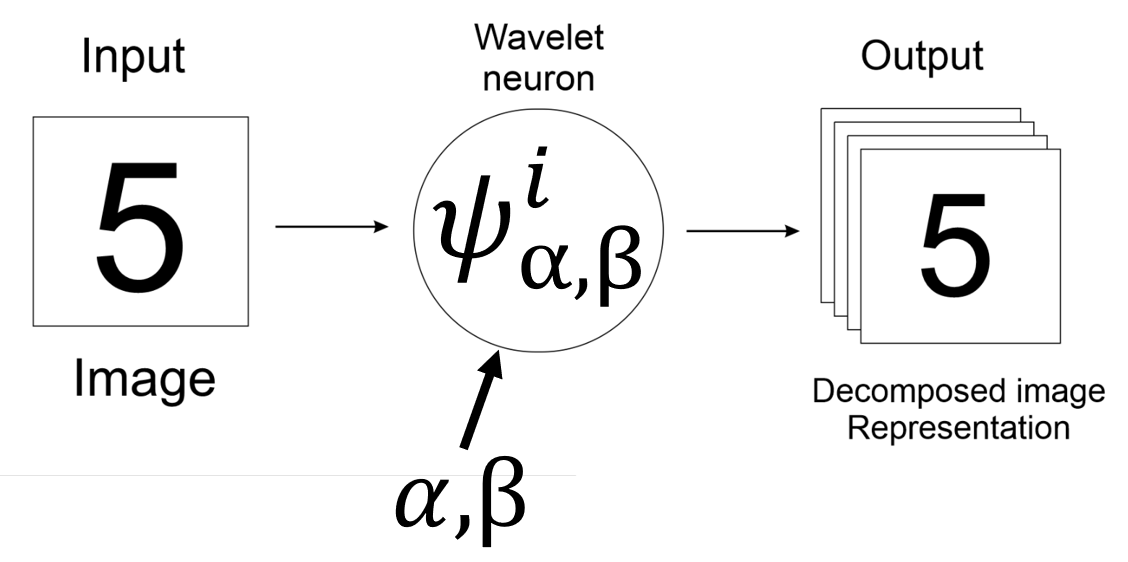}}
		\caption{Input image decomposition of a wavelet neuron that is incorporated with the learnable parameter $\alpha$ and $\beta$.}
		\label{wavelet_neuron}
	\end{center}
\end{figure}

The parameterization in our proposed methodology is based on the defined $\alpha$ and $\beta$. By changing the values of $\alpha$ and $\beta$, one can develop different decomposed image representations. Therefore, we introduce wavelet neurons that perform the wavelet transform of a given input image and is incorporated with the learnable parameters $\alpha$ and $\beta$ (see Figure \ref{wavelet_neuron}). Similar to conventional neural networks, the introduced parameters are implemented as a learnable parameter. During the training phase, the values of $\alpha$ and $\beta$ will be updated using the popular back propagation algorithm.

\subsection{Multi-Path Network Architecture}

\begin{figure*}
	\begin{center}
		\centerline{\includegraphics[width=120mm]{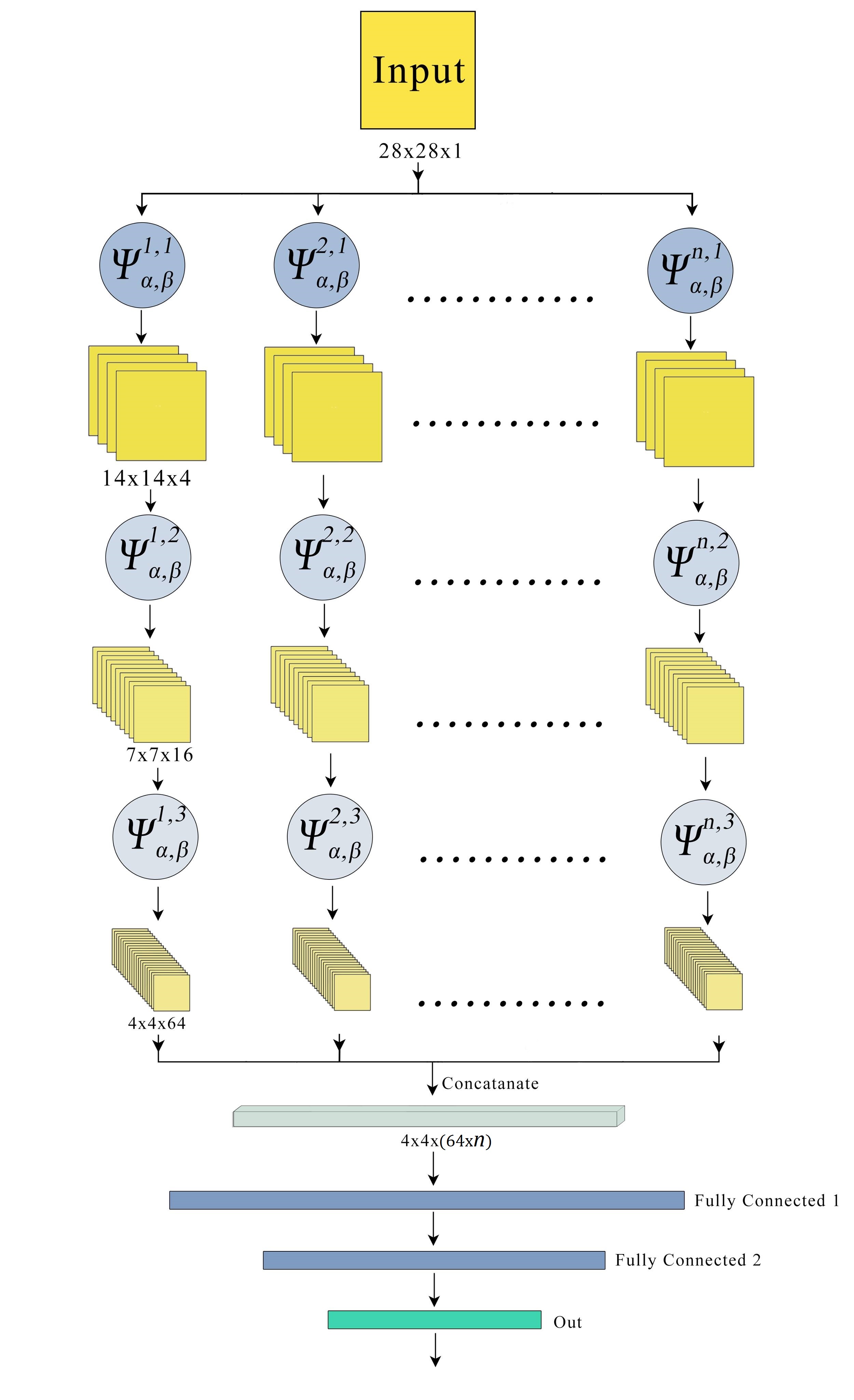}}
		\caption{Wavelet neural network with $n$ decomposition paths with three decomposition levels followed by two fully connected layers.}
		\label{whole_wavelet_network}
	\end{center}
\end{figure*}

In this work, we propose a novel network architecture that has multiple wavelet decomposition paths containing wavelet neurons as hidden units followed by fully connected layers. Each wavelet neuron is parameterized by two trainable variables, which takes an input image or a feature map of dimension  $n \times n \times d$ and outputs a decomposition of dimensions $n/2 \times n/2 \times 4d$ corresponding to three detail coefficients and one approximation coefficient as previously explained.(see Figure \ref{wavelet_neuron}). Wavelet neurons can be arranged in a sequential manner where it is possible to feed the decompositions from the first neuron to the second and so on to perform hierarchical levels of decompositions from original to coarse level. In this work we call such a sequential set of wavelet neurons as a Path. The architecture in Figure \ref{whole_wavelet_network} depicts $n$ number of wavelet paths stacked together where each path contains $3$ wavelet neurons. Therefore, each path will contain six trainable parameters. The output of the wavelet paths are concatenated then flatted and fed to fully connected layers. In our work, we have used two fully connected layers with $32$ conventional neurons per each. We have developed several different architecture configurations by changing the number of parallel wavelet paths to yield networks with different complexities, which led us to analyze the performance and learning of the introduced wavelet parameters. Hence, we have chosen the best performed architecture and compared it against the baseline deep learning models. Table \ref{table_archi_confg} shows the developed architecture configuration with naive classification results on a preliminary dataset (MNIST without data augmentation).

\begin{table}[h]
	\caption{Performance of several proposed wavelet CNN architecture configurations (trained on MNIST).}
	\centering
	\begin{tabular}{|l|l|l|l|l|}
		\hline
		Wavelet &   Wavelet     &    Fully 		& Total       & Error      \\
		Paths &   Parameters  &    Connected  &  Parameters & Rate \%    \\
				&   			&    Layers     &  			  &  		   \\
		\hline
		1 & 6  &  1 & 32k & 1.79\\
		1 & 6  &  2 & 32k & 1.28\\
		2 & 12 &  2 & 66k & 1.14\\
		4 & 24 &  2 & 132k & 1.45\\
		6 & 36 &  2 & 198k & 1.09\\
		\textbf{8} & \textbf{48} &  \textbf{2} & \textbf{264k} & \textbf{0.89}\\
		16 & 96 &  2 & 525.7k & 1.69\\
		32 & 192 &  2 & 1.1M & 2.14\\
		\hline
	\end{tabular}\label{table_archi_confg}
\end{table}

Firstly, the input images to the model are preprocessed by normalizing and then passed through the wavelet paths. There, these inputs are subjected to three distinct wavelet decomposition levels. At the end of the multi-path layout, after three levels of decompositions, feature maps are concatenated and connected with fully connected layers. Rectified Linear units (ReLu) have been used as the activation function for the fully connected layers and sigmoid activation function has been used for the wavelet neurons. During the training process, we used dropout \cite{cite_37} with $0.8$ of keep probability for fully connected layers to prevent over-fitting.

\section{Evaluation}

\subsection{Experiment}

In this section, we discuss the evaluation of the proposed wavelet neural network for image classification. We train and test the system on publicly available image datasets: MNIST ~\cite{cite_32}, CIFAR-10 \cite{cite_41}, CIFAR-100 \cite{cite_41} and SVHN \cite{cite_42}. MNIST is a popular and one of the most common preliminary datasets in machine learning practice. The dataset composed of $28$ by $28$ gray-scale handwritten digit images from $0-9$ with $60,000$ training data and $10,000$ testing data. CIFAR-10 and CIFAR-100 contain $32$ by $32$ color images with $50,000$ training images and 10,000 testing images, with 10 and 100 distinct classes respectively. SVHN contains $32$ by $32$ color images of house numbers from Google Street View under 10 classes (digits), which includes 73,257 and 26,032 examples for training and testing respectively. The experiments for MNIST, CIFAR-10 and CIFAR-100 were conducted without any data augmentation.

We have used Nvidia Jetson TX2 development kit as the experimental hardware. The implementation of the wavelet neural network has been done using Tensorflow framework. Moreover, the Adam optimizer \cite{cite_33} has been used as the optimizer with exponential learning rate decay initialized from $0.01$. Then we repeat the learning process $10$ times to obtain the average classification accuracy. In order to compare our findings, we have selected several influential deep learning models that are also trained and tested against the same datasets. These networks are selected based on the number of parameters and the depth of the networks including the state of the art.


\subsection{Results and Discussion}
\label{results}

We have compared the performance of the proposed networks in terms of accuracy and number of parameters under above mentioned datasets. According to the performance evaluation reported in Table \ref{table_archi_confg}, we have selected the architecture configuration with $8$ parallel paths since it gave the best performance. Then we have conducted further experiments using this configuration and the results were compared against the results of influential deep learning models. Moreover, we have compared our findings with several previous work done in the wavelets and neural networks domain.

\subsubsection{MNIST:}

\begin{table}[h]
	\centering
	\caption{Comparison between the proposed best performing architecture and the baseline deep learning models for MNIST dataset.}\label{table_mnist}
	\begin{tabular}{|l|l|l|}
		\hline
		Model & \# Parameters & Error rate \% \\
		\hline
		\textbf{Learnable Wavelet Neural Network} &  \textbf{264k} & \textbf{0.27}\\
		AlexNet \cite{cite_1} & 61M & 0.44\\
		VGG16 \cite{cite_2} & 138M & 0.37\\
		Inception V3 \cite{cite_34} & 23M & 0.33\\
		ResNet 152 \cite{cite_25}  & 60M & 0.30\\
		DropConnect \cite{cite_36} (State of the art) & 2.5M & 0.21\\
		Network in Network \cite{cite_43} & 1M & 0.43\\
		Max-out network \cite{cite_45} & 6M & 0.39\\
		Wavelet network approach--Mexican hat wavelet \cite{cite_48} & -- & 0.8\\
		Wavelet network approach--Morlet wavelet \cite{cite_48} & -- & 0.79\\
		Wavelet network approach--RASP wavelet \cite{cite_48} & -- & 0.8\\
		MMEE-AlexNet \cite{cite_31} & 61M & 0.49\\
		NEE-AlexNet \cite{cite_31} & 61M & 0.57\\
		\hline
	\end{tabular}
\end{table}

We tested our architecture without any data augmentation under MNIST dataset and achieved near state of the art results by training a lesser number of parameters compared to other models. Table \ref{table_mnist} shows a comparison under accuracy and number of parameters between chosen models starting from shallow to deep. It is evident that our network has the least number of parameters. Compared to DropConnect we have achieved a significant parameter reduction and that is approximately $90\%$. However, we were able to achieve an error rate of $0.29\%$, which is the second best error rate and only lagging behind the DropConnect architecture. Although, deeper networks like VGG16 has achieved a $0.37\%$ error it contains $523$ times more parameters than our network. Even though our model does not achieve the state of the art classification results, this comparison clearly evidence that our architecture is able to reach a highly acceptable accuracy level with very low number of parameters. Further comparisons with the previous work \cite{cite_48} and \cite{cite_31} have also shown in the table. It has been used different wavelet basis functions and developed a wavelet network to extract features. However, our model has outperformed this work by considerable margins.

\begin{table}[h]
	\centering
	\caption{Comparison between the baseline deep learning models and our best performed model for CIFAR-10 and CIFAR-100 datasets in terms of number of parameters and test accuracy.}\label{table_cifar}
	\begin{tabular}{|l|l|l|l|}
		\hline
		Model & \# Parameters & CIFAR-10 & CIFAR-100\\
		\hline
		\textbf{Learnable Wavelet Neural Network} &  \textbf{264K} & \textbf{94.87\%} & \textbf{81.22\%} \\
		AlexNet \cite{cite_1} & 61M & 94.19\% & 80.74\% \\
		DenseNet \cite{cite_42} & 27.2M & 93.95\% & 80.75\%\\
		VGG16 \cite{cite_2} & 138M & 92.45 & 77.28\% \\
		Inception V3 \cite{cite_34} & 23M & 92.16\% & 76.49\% \\
		ResNet-110 \cite{cite_25} & 1.7M & 91.57\% & 74.84\% \\
		Network in Network \cite{cite_43} & 1M & 90.89\% & 73.32\% \\
		Max-out network \cite{cite_45}& 6M & 92.32\% & 76.46\% \\
		FitNet \cite{cite_44}  & 1M & 94.61\% & 81.06\% \\
		LeNet \cite{cite_46}  & 60K & 73.3\% & 41.1\% \\
		WaveLeNet \cite{cite_18} & 60K & 72.4\% & 39.6\% \\
		CNN--WAV 2 \cite{cite_47} & -- & 76.42\% & -- \\
		CNN--WAV 4 \cite{cite_47} & -- & 85.67\% & -- \\
		MMEE-AlexNet \cite{cite_31} & 61M & 91.15\% & --\\
		NEE-AlexNet \cite{cite_31} & 61M & 90.62\%  & --\\
		\hline
	\end{tabular}
\end{table}

\newpage
\subsubsection{CIFAR-10/CIFAR-100:}Table \ref{table_cifar} elaborates on the results of both CIFAR-10 and CIFAR-100. We have achieved a 94.87\% accuracy without any data augmentation. And our model contains the least number of trainable parameters among all the models in comparison. For CIFAR 100 our model have achieved an accuracy of 81.22\% but still holds its position as the model that contains the least number of parameters. FitNet, who has reported the second best performance has twice the number of parameters compared to our model. Moreover, results from one of the previous work \cite{cite_18} has compared, which is also outperformed by the proposed network despite the number of parameters. Results from another wavelets and CNN classification work \cite{cite_47} and wavelet based preprocessing for CNN \cite{cite_31} have also compared. Both previous work process inputs in wavelet domain using fixed wavelets without any wavelet learning.  

\begin{table}[h]
	\centering
	\caption{Comparisons of performance by error rate on SVHN dataset.}\label{table_svhn}
	\begin{tabular}{|l|l|}
		\hline
		Model & Accuracy \% \\
		\hline
		\textbf{Learnable Wavelet Neural Network} & \textbf{97.51} \\
		VGG16 \cite{cite_2} & 96.86\\
		DenseNet \cite{cite_42} & 97.13 \\
		ResNet 152 \cite{cite_25}  & 96.72 \\
		Network in Network \cite{cite_43} & 96.65\\
		Max-out network \cite{cite_45} & 96.78 \\
		MMEE-AlexNet \cite{cite_31}  & 94.43\\
		NEE-AlexNet \cite{cite_31}  & 94.26\\
		\hline
	\end{tabular}
\end{table}

\subsubsection{SVHN:} For further evaluations, we have included Table \ref{table_svhn} to show the accuracy performance for SVHN dataset with simple data-augmentation including shifting, rotating and color inverting. Our network again beats all the compared models and the closest margin is 0.38\% with the DenseNet.

\section{Conclusion}
\label{conclusion}

We have presented a novel wavelet neural network architecture with learnable wavelet parameters. The specialty is that our network contains the least number of trainable parameters compared to conventional and concurrent deep learning models. We have tested several different architecture configurations to select the best performing network. Then we tested this particular configuration on several common image datasets, and compared the classification results and number of parameters against the performance of influential deep learning models. The experimental results display that our model has been able to outperform all the compared deep learning models while training only $264$K parameters without any data augmentation. The reconstruction of original inputs from the developed feature maps have shown slight distortion because of the sigmoid activation functions. 

However, wavelet based formulation and decomposition consume a significant amount of computational cost and makes it difficult to train, which can be introduced as limitations. The implementations and operations have to be optimized to perform the calculations within lesser cost. Therefore, we would like to introduce this as a preliminary work in learnable wavelets with neural networks. For future work, the proposed network need to be extended to classify more complex datasets that includes diverse shapes and patterns. Moreover, the concept of learnable parameterized wavelets can be combined with other deep learning architectures such as CNNs for better feature extraction and learning.


\end{document}